\documentclass[10pt,twocolumn,letterpaper]{article}

\usepackage{iccv}
\usepackage{times}
\usepackage{epsfig}
\usepackage{graphicx}
\usepackage{amsmath}
\usepackage{amssymb}
\usepackage{subcaption}
\usepackage{adjustbox}

\usepackage[pagebackref=true,breaklinks=true,letterpaper=true,colorlinks,bookmarks=false]{hyperref}

\iccvfinalcopy 

\begin{document}

\title{CLRerNet: Improving Confidence of Lane Detection with LaneIoU}

\author{Hiroto Honda\\
GO Inc. \\
{\tt\small hiroto.honda@goinc.jp}
\and
Yusuke Uchida\\
GO Inc. \\
{\tt\small yusuke.uchida@goinc.jp}
}

\maketitle

\begin{abstract}
Lane marker detection is a crucial component of the autonomous driving and driver assistance systems.
Modern deep lane detection methods with row-based lane representation exhibit excellent performance on lane detection benchmarks.
Through preliminary oracle experiments, we firstly disentangle the lane representation components to determine the direction of our approach.
We show that correct lane positions are already among the predictions of an existing row-based detector, and the confidence scores that accurately represent intersection-over-union (IoU) with ground truths are the most beneficial.
Based on the finding, we propose LaneIoU that better correlates with the metric, by taking the local lane angles into consideration.
We develop a novel detector coined CLRerNet featuring LaneIoU for the target assignment cost and loss functions aiming at the improved quality of confidence scores.
Through careful and fair benchmark including cross validation, we demonstrate that CLRerNet outperforms the state-of-the-art by a large margin - enjoying F1 score of 81.43\% compared with 80.47\% of the existing method on CULane, and 86.47\% compared with 86.10\% on CurveLanes. 
Code and models are available at \url{https://github.com/hirotomusiker/CLRerNet}.
\end{abstract}

\begin{figure}[t]
\begin{center}
   \includegraphics[width=0.99\linewidth]{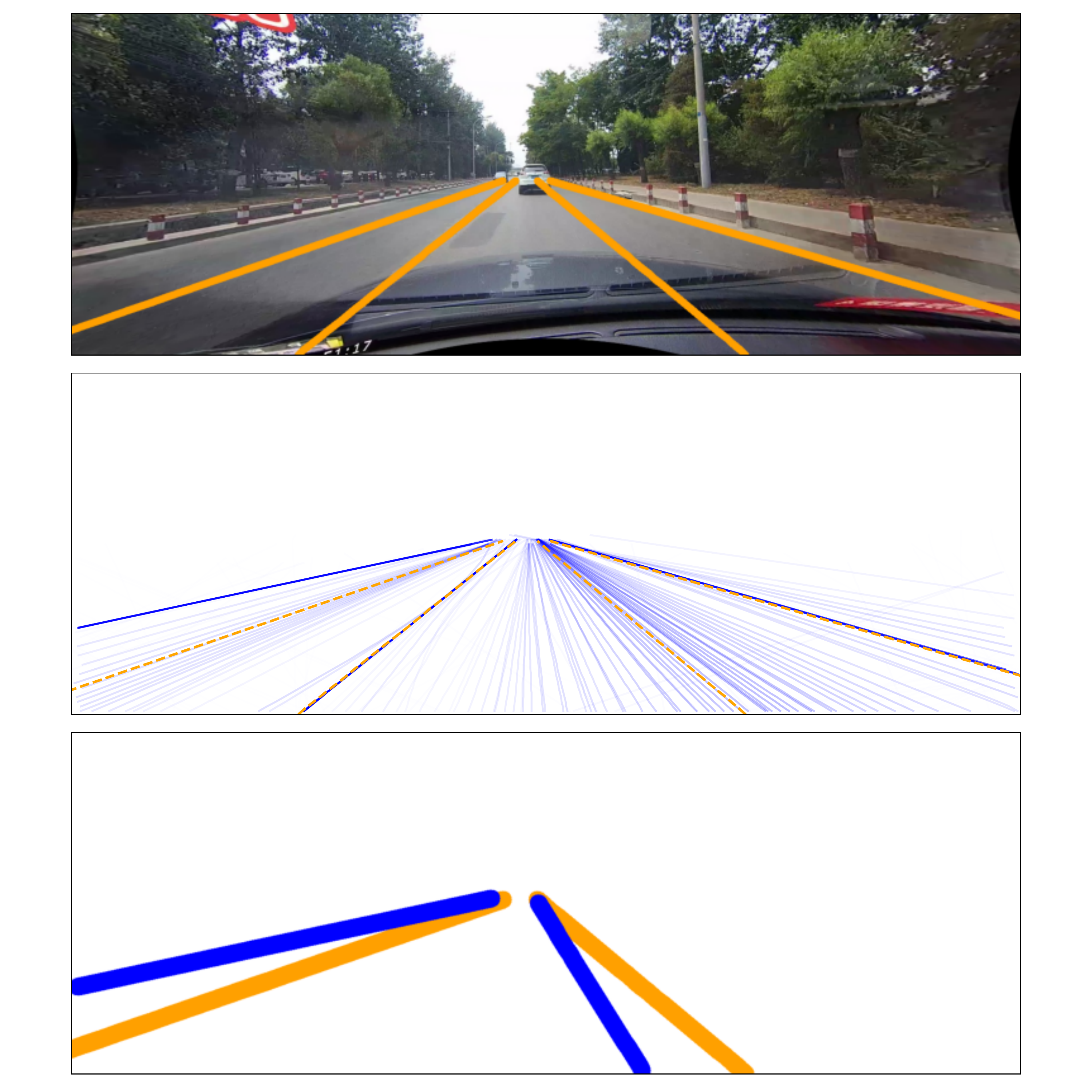}
\end{center}
   \caption{Lane detection example. top: ground truth. middle: all the predictions (blue, deeper for higher confidence scores) and ground truths (dashed orange). bottom: metric IoU calculation by comparing segmentation masks of predictions (blue) and GTs (orange). Best viewed in color.}
\label{fig1}
\end{figure}

\section{Introduction}

Lane (marker) detection plays an important role in the autonomous driving and driver assistance systems.
Like other computer vision tasks, emergence of convolutional neural networks (CNNs) has brought rapid progress on lane detection performance. 
Modern lane detection methods are grouped into four categories in terms of lane instance representation.
Segmentation-based \cite{pan2018SCNN, RESA_2021} and keypoint-based \cite{Qu_2021_CVPR} representations regard lanes as segmentation mask and keypoints respectively.
The parametric representation methods \cite{TorresBPBSO20, LSTR} utilize curve parameters to regress lane shapes. 
The row-based representation \cite{tabelini2021cvpr, Zheng_2022_CVPR, qin2020ultra, qin2022ultrav2, Liu_2021_ICCV} regards a lane as a set of coordinates on the certain horizontal lines. 
The first two representations are employed in the bottom-up detection paradigm, where the lane positions are directly detected in the image and grouped into lane instances afterwards. The latter two are adopted to the top-down instance detection methods where each lane detection is regarded as both global lane instance and a set of local lane points.
The row-based representation is the de-facto standard in terms of detection performance among the above representation types.
We choose the best-performing CLRNet \cite{Zheng_2022_CVPR} from the row-based methods as our baseline.

The performance of lane detection relies on lane point localization and instance-wise classification. 
The lane detection benchmarks \cite{pan2018SCNN, CurveLane-NAS} employ segmentation-mask-based intersection-over-union (IoU) between predicted and ground-truth (GT) lanes as an evaluation metric.
The predicted lanes whose scores are above the predefined threshold are treated as valid predictions to calculate F1 score.
Therefore, the predicted lanes with large segmentation-based IoUs with GTs should have large classification scores. 

To determine the direction of our approach, we firstly conduct the preliminary oracle experiments h by replacing the confidence score, anchor parameters and length of each prediction with the oracle values.
By making the confidence scores oracle, the F1 score goes to near-perfect 98.47\%.
The result implies the correct lanes are already among the predictions, however the confidence scores need to be predicted accurately representing the metric IoU.
Fig. \ref{fig1} (middle) shows the comparison between all the predictions (blue) and GTs (dashed orange). The color depth of the predictions is proportional to their confidence scores. 
The left-most prediction is the high-confidence false positive that misses the ground truth, however there is a correct low-confidence prediction near the GT which has a high IoU.

The next question is: how could the segmentation-based IoU be implemented as a learning target?
In the row-based methods, the prediction and GT lanes are both represented as sets of $x$ coordinates at the fixed rows. 
\cite{Zheng_2022_CVPR} introduces the LineIoU loss to measure the intersection and union row by row and sum them up respectively.
However, this approach is not equivalent to the segmentation-based IoU especially for the non-vertical, tilted lanes (e.g. Fig. \ref{fig1} bottom) or curves.
We introduce the novel IoU coined \textit{LaneIoU}, which takes local lane angles of the lanes into account.
The LaneIoU integrates the angle-aware intersection and union of each row to match the segmentation-based IoU.

The row-based methods learn global lane probability scores for each anchor.
The dynamic sample assignment employed in the recent object detector \cite{ota, YOLOX} is also effective for lane detection training \cite{Zheng_2022_CVPR}. The IoU matrix and the cost matrix between the predicted lanes and the GTs respectively determine the number of anchors to assign for each GT and which anchors to assign.
The confidence targets of the assigned anchors are set to positive (one). Therefore, sample assignment is responsible for learning the confidence scores.
We introduce LaneIoU to sample assignment in order to bring the detector's confidence scores close to the segmentation-based IoU.  LaneIoU dynamically determines the number of anchors to assign and prioritizes the anchors to assign as a cost function.
Moreover, the loU loss to regress the horizontal coordinates is also replaced by our LaneIoU to appropriately penalize the predicted lanes at different tilt angles.
The LaneIoU integration to CLRNet \cite{Zheng_2022_CVPR} makes the detector's training more straightforward, thus we coin our method \textit{CLRerNet}.

We showcase the effectiveness of LaneIoU through extensive experiments on CULane and CurveLanes and report the state-of-the-art results on both datasets.
Importantly, for a reliable and fair benchmark, we employ the average score of five models for each experiment condition, while prior work shows a score of a single model.
Moreover, the F1 metric employed in the lane detection evaluation is utterly sensitive to the detector's lane confidence threshold, thus we determine the threshold utilizing the 5-fold cross validation on the train split.

Our contributions in this paper are threefold:
\begin{itemize}
	\item \textit{Clearer focus}: Through preliminary oracle experiments, we show that correct lane positions are already among the predictions of an existing detector, and the confidence scores that represent intersection-over-union (IoU) with ground truths are the most effective to improve performance.
	\item \textit{Clearer training method}: As a lane similarity function, we leverage LaneIoU which well correlates with the evaluation metric and integrate it into training as a sample assignment cost and regression target.
	\item \textit{Clearer benchmarking}: Multi-model evaluation and cross-validation-based score thresholding are employed for fair benchmark. 
The effectiveness and generality of LaneIoU is verified and CLRerNet achieves state-of-the-art in the CULane and CurveLanes benchmarks.
\end{itemize}

\section{Related Work}
\label{sec:relatedwork}

\subsection{Object detection}
\noindent{\textbf{Training sample assignment.}} Sample assignment is the major research focus in object detection. The proposals from the detection head are assigned to the ground truth samples. \cite{fasterrcnn, fpn, maskrcnn, retinanet} assign the GTs by calculating IoU between the anchors on the feature map grid and GT boxes statically. \cite{ota} introduces the optimal transfer assignment (OTA) for object detector's training sample assignment, that dynamically assigns the prediction boxes to the GTs. \cite{YOLOX} simplifies OTA and realizes iteration-free assignment. 

\noindent{\textbf{IoU functions.}} 
Several variants of IoU functions \cite{Rezatofighi_2018_CVPR, diou, ciou} are proposed for accurate bounding box regression and fast convergence. For example, the generalized IoU (GIoU) \cite{Rezatofighi_2018_CVPR} introduces the smallest convex hull of the boxes and makes IoU differentiable even when the bounding boxes do not overlap. Our LaneIoU is based on GIoU but newly enables the IoU calculation between curves in the row-based representation.

\subsection{Lane detection}
Lane detection paradigms are grouped by lane representation types, namely segmentation-based, keypoint-based, row-based and parametric representations.

\noindent{\textbf{Segmentation-based representation.}}  This line of work is the bottom-up pixel-based estimation of lane existence probability.
SCNN \cite{pan2018SCNN} and RESA \cite{RESA_2021} employ a semantic segmentation paradigm to classify the lane instances as separate classes on each pixel.
The correspondence between lane and class is determined by annotation thus not flexible (e.g. some lane position may belong to two classes). 
The benchmark datasets \cite{pan2018SCNN, CurveLane-NAS} employ pixel-level IoU to compare predicted lanes with GTs, and are friendly to the segmentation-based methods.
However the methods do not treat lanes as holistic instances and require post-processing which is computationally costly.
\cite{qin2020ultra, Zheng_2022_CVPR} exploit the segmentation task as the auxiliary loss only during training time to improve the backbone network. We follow these methods and adopt the auxiliary branch and loss.

\noindent{\textbf{Keypoint-based representation.}} Similar to human pose estimation, the lane points are detected as keypoints and grouped into lane instances afterwards.
PINet \cite{pinet} employs test-time detachable stacked hourglass networks to learn keypoint probabilities and cluster the keypoints into the lane instances. 
FOLOLane \cite{Qu_2021_CVPR} also detects lanes as keypoints inspired by the bottom-up human pose detection method \cite{openpose}.
GANet \cite{ganet-cvpr2022} regresses the offsets of the detected keypoints from the starting point of the corresponding lane instances.
This line of methods requires post-process to group the lane points into lane instances, which is computationally expensive.

\noindent{\textbf{Parametric representation.}} In this line of work, a lane instance is represented as a set of curve parameters. 
PolyLaneNet \cite{TorresBPBSO20} employs a curve representation using polynomial coefficients.
LSTR \cite{LSTR} employs end-to-end transformer-based lane parameter set detection.
BSNet \cite{BSNet} chooses the quasi-uniform b-spline curves and shows the highest F1 score among this category.
These methods achieve relatively fast inference, however an error of one parameter holistically affects the lane shape.

\noindent{\textbf{Row-based representation.}} The lane instance is represented as a set of x-coordinates at the fixed rows. LaneATT \cite{tabelini2021cvpr} employs lane anchors to learn the confidence score and local x-coordinate displacement for each anchor. An anchor is defined as a fixed angle and a start point. The training target is assigned statically according to the horizontal distance between each anchor to GTs. CLRNet \cite{Zheng_2022_CVPR} adopts learnable anchor parameters (start point $x_a$, $y_a$ and $\theta_a$) and length $l$.  For sample assignment, the simplified optimal transport assignment \cite{YOLOX} is employed to dynamically allocate the closest predictions to each ground truth. 
Both methods pool the feature map by the anchors and feed the extracted features to the head network. The head network outputs the classification and regression tensors for each anchor. This paradigm corresponds to the 2-stage object detection methods such as \cite{fpn, maskrcnn}
UFLD \cite{qin2020ultra} captures the global features by flattening the feature map and learns row-wise lane position classification. UFLDv2 \cite{qin2022ultrav2} extends \cite{qin2020ultra} to a row- and column-wise lane representation to deal with the near-horizontal lanes. 
CondLaneNet \cite{Liu_2021_ICCV} learns a probability heatmap of lane start points from where the dynamic convolution kernels are extracted. The dynamic convolution is applied to the feature map, whereby the row-wise lane point classification and x-coordinate regression are carried out.
LaneFormer \cite{laneformer2022} employs a transformer with row and column attention to detect lane instances in an end-to-end manner. Additionally vehicle detection results are fed to the decoder to make the pipeline object-aware. The row-based representation is the de-facto standard in terms of detection performance among the four representation types. 

\section{Methods}
\subsection{Network design and losses}
\begin{figure*}[t]
\begin{center}
    \includegraphics[width=.999\linewidth]{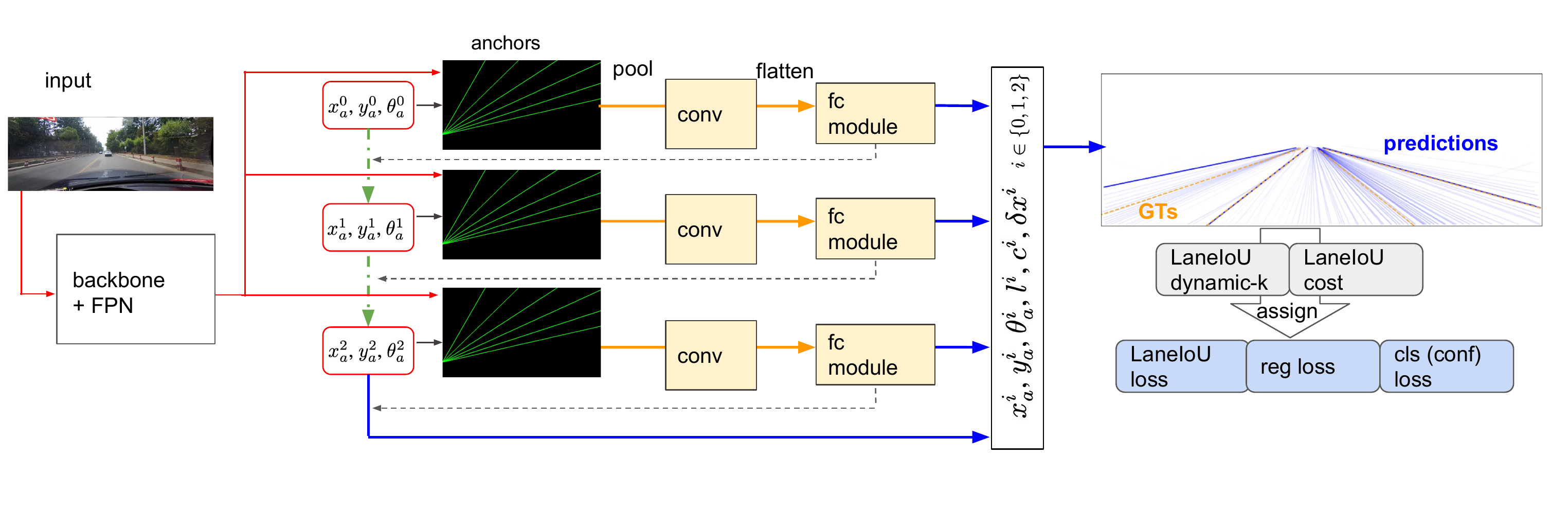}
\end{center}
   \caption{Network Schematic of CLRerNet.}
\label{fig:long}
\label{fig:schematic}
\end{figure*}

The row-based representation \cite{Zheng_2022_CVPR, Liu_2021_ICCV, tabelini2021cvpr} leverages the most accurate but simple detection pipeline among the four types described in Section \ref{sec:relatedwork}. From the row-based methods we employ the best-performing CLRNet \cite{Zheng_2022_CVPR} as a baseline. 
The network schematic is shown in Fig. \ref{fig:schematic}.
The backbone network (e.g. ResNet \cite{He_2016_CVPR} and DLA \cite{Yu_2018_CVPR}) and the up-sampling network extract multi-level feature maps whose spatial dimensions are (1/8, 1/16, 1/32) of the input image.
The initial anchors are formed from the $N_a$ learnable anchor parameters $(x_a, y_a, \theta_a)$ where ($x_a$, $y_a$) is the starting point and $\theta_a$ the tilt of the anchor.
The feature map is sampled along each of them and fed to the convolution and fully-connected (FC) layers.
The classification logits $c$, anchor refinement $\delta x_a$, $\delta y_a$, $\delta \theta_a$, length $l$ and local x-coordinate refinement $\delta x$ tensors are output from the FC layers. 
The anchors refined by $\delta x_a$, $\delta y_a$ and $\delta \theta_a$ re-sample the higher-resolution feature map, and the procedure is repeated for three times.
The pooled features are interacted with the feature map via cross-attention and are concatenated across different refinement stages.
The lane prediction is expressed as classification (confidence) logits and a set of x-coodinates at $N_{row}$ rows calculated from the final $x_a$, $y_a,$ $\theta_a$, $l$ and $\delta x$.
More details about the refinement mechanism can be found in \cite{Zheng_2022_CVPR}.
During training, the predictions close to a GT are assigned via dynamic assigner \cite{YOLOX}.
The assigned predictions are regressed toward the corresponding GT and learned to be classified as positives. 
\begin{equation}
\label{eq:lossses}
L= \lambda_0 L_{reg} + \lambda_1 L_{cls} + \lambda_2 L_{seg} + \lambda_3 L_{LaneIoU}
\end{equation}
where $L_{reg}$ is smooth-L1 loss to regress the anchor parameters $(x_a, y_a, \theta_a)$ and $l$, $L_{cls}$ a focal loss \cite{focalloss} for positive-or-negative anchor classification, $L_{seg}$ an auxiliary cross-entropy loss for per-pixel segmentation mask, and  $L_{LaneIoU}$ the newly introduced LaneIoU loss. 

\subsection{Oracle experiments}\label{subsection:oracle}

We conduct the preliminary oracle experiments to determine the direction of our approach. The prediction components from the trained baseline model are replaced with GTs partially to analyze how much room for improvement each lane representation component has.
Table \ref{table:oracles} shows the oracle experiment results. The baseline model (first row) is CLRNet-DLA34 trained without the redundant frames. The confidence threshold is set to 0.39 which is obtained via cross-validation (see Subsection \ref{subsec:datasets} and \ref{subsec:train_eval}).

Next, we calculate the metric IoU between predictions and GTs as the oracle confidence scores. For each prediction, the maximum IoU among the GTs is employed as the oracle score. In this case, the predicted lane coordinates are not changed. The $\text{F1}_{50}$ jumps to 98.47 - the near-perfect score (second row). The result suggests that \textbf{the correct lanes are already among the predictions, however the confidence scores need to be predicted accurately representing the metric IoU}.

The other components are the anchor parameters - $x_a$, $y_a$, $\theta_a$ and length $l$ that determine the lane coordinates. We alter the anchor parameters and lane length by those of GTs (third and fourth rows respectively). Although the row-wise refinement $\delta x$ is not changed, the oracle anchor parameters improve $\text{F1}_{50}$ by 9 points. On the other hand, the oracle length does not affect the performance significantly. The results lead to the second suggestion that the anchor parameters ($x_a$, $y_a$, $\theta_a$) are important in terms of lane localization. 

We focus on the first finding and aim to learn high-quality confidence scores by improving the lane similarity function.

\begin{table}
\begin{center}
\begin{tabular}{ccc|c} 
\hline
confidence & ($x_a$, $y_a$, $\theta_a$) & $l$ &$\text{F1}_{50}$   \\
\hline
& & & 80.86 \\
 \checkmark & & & \textbf{98.47} \\
 & \checkmark &  & 89.91  \\
 &  & \checkmark &  81.09 \\
\hline
\end{tabular}
\end{center}
\caption{Oracle experiment results.}
\label{table:oracles}
\end{table}

\subsection{LaneIoU}

\begin{figure}[t]
\begin{center}
    \centering
    \includegraphics[width=.99\linewidth]{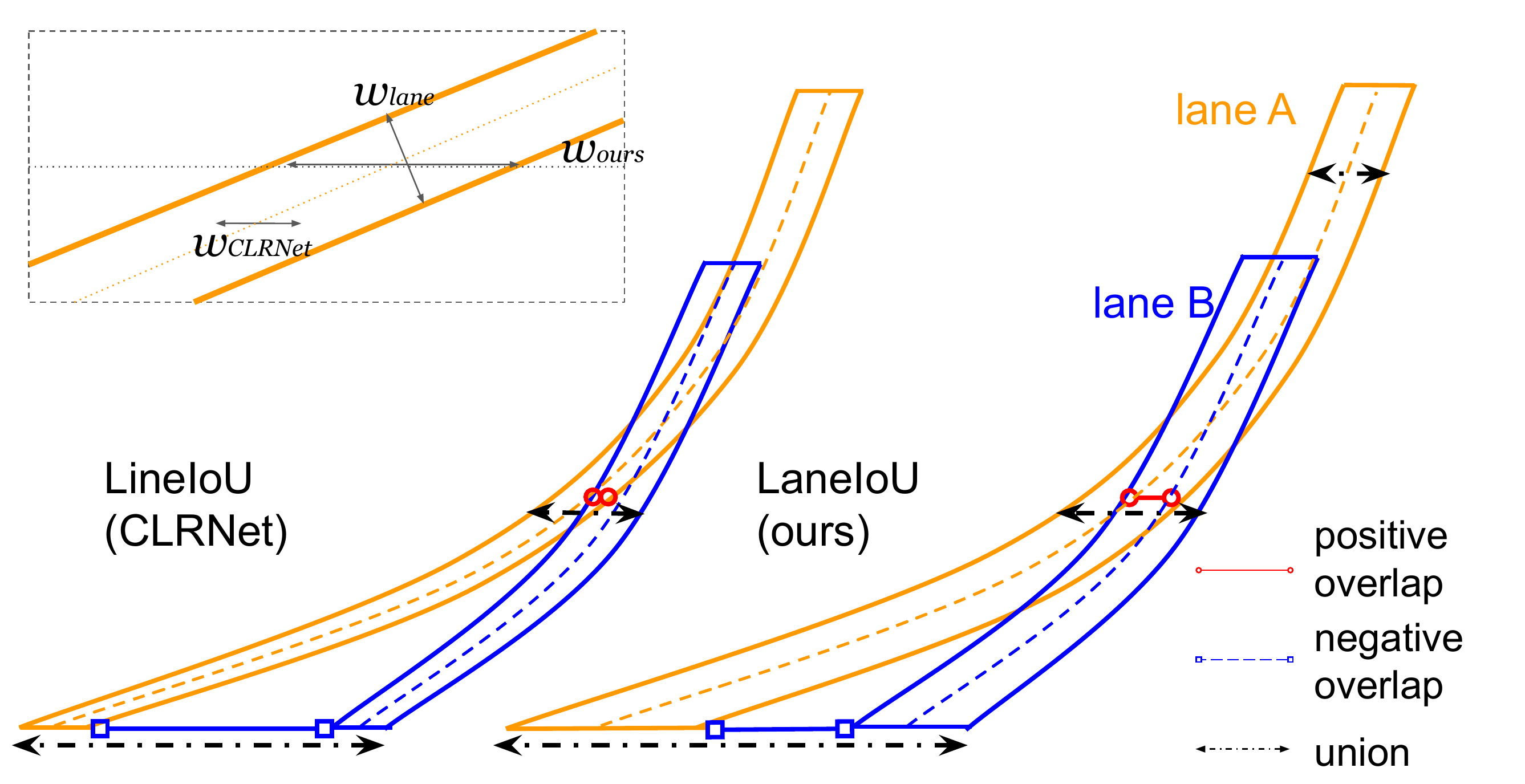}
\end{center}
   \caption{Example of LineIoU \cite{Zheng_2022_CVPR} and LaneIoU calculations between laneA and laneB. $w_{lane}$, $w_{CLRNet}$ and $w_{ours}$ within the dashed rectangle stand for the lane width, the constant width of \cite{Zheng_2022_CVPR} and our angle-aware width.}
\label{fig:laneiou}
\end{figure}

The existing methods \cite{Liu_2021_ICCV} and \cite{Zheng_2022_CVPR} exploit horizontal distance and horizontal IoU as similarity functions respectively. However, these definitions do not match the metric IoU calculated with segmentation masks. 
For instance, when the lanes are tilted, the horizontal distance corresponding to the certain metric-IoU is larger than that of vertical lanes.
To bridge the gap, we introduce a differentiable local-angle-aware IoU definition, namely LaneIoU. 
Fig. \ref{fig:laneiou} shows an example of IoU calculation between two tilted curves. We compare LineIoU \cite{Zheng_2022_CVPR} and LaneIoU on the same lane instance pair. 
LineIoU applies a constant virtual width regardless of the lane angle and the virtual lane gets 'thin' at the tilted part.
In our LaneIoU, overlap and union are calculated considering the tilt of the local lane parts at each row. We define $\Omega_{pq}$ as the set of y slices where both lanes $p$ and $q$ exist, and $\Omega_{p}$ or $\Omega_{q}$ where only one lane exists. LaneIoU is calculated as:
\begin{equation}
\label{laneiou}
LaneIoU=\frac{\sum_{i=0}^H I_i }{\sum_{i=0}^H U_i}
\end{equation}
where $I_i$ and $U_i$ are defined as:
\begin{equation}
\label{laneiou}
I_i=\text{min}(x_i^p+w_i^p, x_i^q+w_i^q)-\text{max}(x_i^p-w_i^p, x_i^q-w_i^q)
\end{equation}
\begin{equation}
\label{laneiou}
U_i=\text{max}(x_i^p+w_i^p, x_i^q+w_i^q)-\text{min}(x_i^p-w_i^p, x_i^q-w_i^q)
\end{equation}
when $i \in\Omega_{pq}$. The intersection of the lanes is positive when the lanes are overlapped and negative otherwise.

\noindent If  $i \notin\Omega_{pq}$, $I_i$ and $U_i$ are calculated as follows:
\begin{equation}
I_i=0,   
U_i=2w_i^k \;\; \text{if} \; k\in\{p, q\}, i \in\Omega_{k}
\end{equation}
\begin{equation}
I_i=0,    
U_i=0 \;\; \text{if} \; i\notin(\Omega_{pq}\cup\Omega_{p}\cup\Omega_{q})
\end{equation}
The virtual lane widths $w_i^p$ and $w_i^q$ are calculated taking the local angles into consideration: 
\begin{equation}
\label{eq:length}
w_i^k = \frac{w_{lane}}{2}\frac{\sqrt{(\Delta x_i^k)^2 +  (\Delta y_i^k)^2}}{\Delta y_i^k}
\end{equation}
where $k\in\{p, q\}$ and $\Delta x_i$ and $\Delta y_i$ stand for the local changes of the lane point coordinates. Equation \ref{eq:length} compensates the tilt variation of lanes and represents a general row-wise lane IoU calculation. When  the lanes are vertical, $w_i$ equals to $w_{lane} / 2$ and gets larger as the lanes tilt. 
$w_{lane}$ is the parameter which controls the strictness of the IoU calculation.
The CULane metric employs 30 pixels for the resolution of (590, 1640).

\begin{figure}[htb]
\centering
  \begin{subfigure}[b]{.495\linewidth}
    \centering
    \includegraphics[width=.999\linewidth]{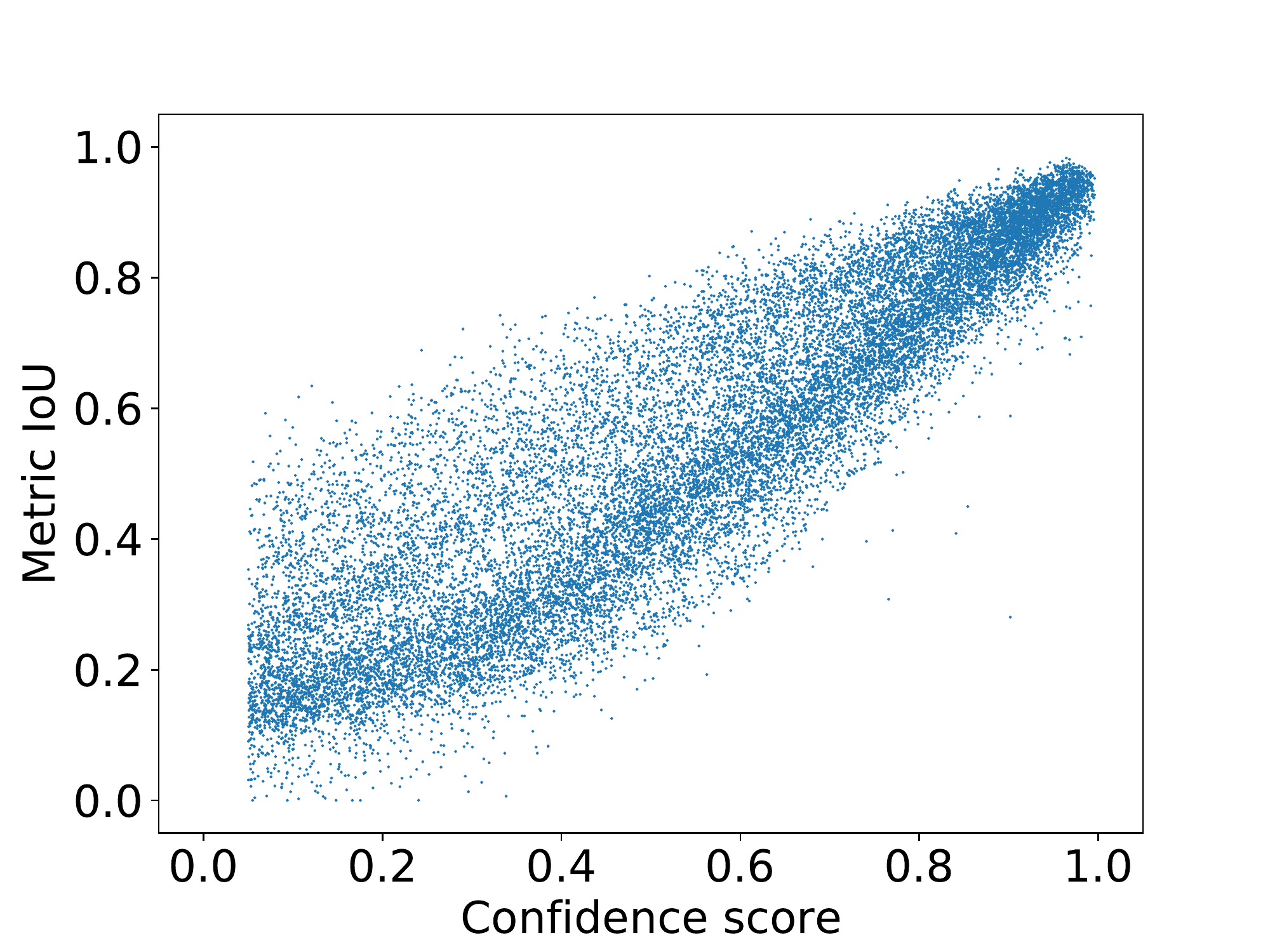}
    \caption{}
  \end{subfigure}
  \begin{subfigure}[b]{.495\linewidth}
    \centering
    \includegraphics[width=.999\linewidth]{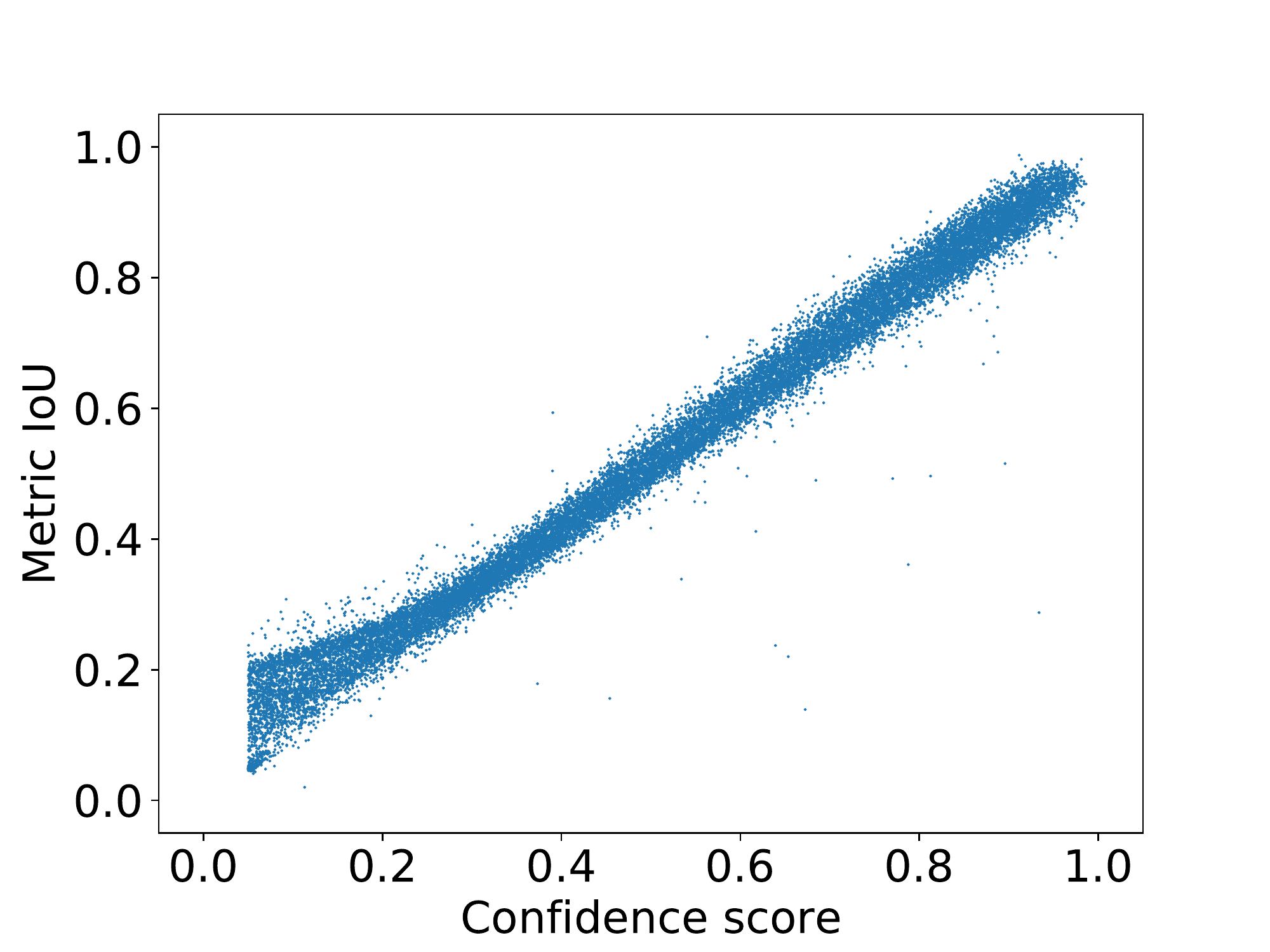}
    \caption{}
  \end{subfigure}\\%
   \caption{Correlation between CULane metric IoU and (a) LineIoU \cite{Zheng_2022_CVPR} and (b) LaneIoU (ours).}
\label{fig:correlation}
\end{figure}

In Fig. \ref{fig:correlation}, LineIoU \cite{Zheng_2022_CVPR} and our LaneIoU are compared by calculating correlation with the CULane metric.
We replace each prediction's confidence score with the LineIoU or LaneIoU value and also calculate the metric IoU. The GT with the largest IoU is chosen for each prediction.
Clearly our LaneIoU shows better correlation with the metric IoU mainly as the result of eliminating the influence of lane angles.

\subsection{Sample assignment}
The confidence scores are learned to be high if the anchor is assigned as positive during training. We adopt LaneIoU for sample assignment to bring the detector's confidence scores close to the segmentation-based IoU.
\cite{Zheng_2022_CVPR} employs the SimOTA assigner \cite{YOLOX} to dynamically assign $k_i$ anchors for each GT lane $t_i$. The number of anchors $k_i$ is determined by calculating the sum of all the anchors' positive IoUs. We employ LaneIoU as:
\begin{equation}
\label{eq:dynamick}
k_i=\sum_{j=1}^mLaneIoU(p_j, t_i)
\end{equation}
where $m$ is the number of anchors and $i=0, 1, ..., n$ is the index of $n$ GT lanes. $p_j$ is a predicted lane and $j=0, 1, ..., m$ is the index of $m$ predictions.
$k_i$ is clipped to be from 1 to $k_{max}$.
The cost matrix determines the priority of the assignment for each GT.
In object detection, \cite{YOLOX} adopts the sum of bounding box IoU and classification costs.
In lane detection, CLRNet \cite{Zheng_2022_CVPR} utilizes the classification cost and the lane similarity cost which consists of horizontal distance, angle difference and start-point distance.
However, the cost that directly represents the evaluation metric is more straightforward for prioritizing predictions.
We define the cost matrix as:
\begin{equation}
\label{eq:cost}
cost_{ji}=-LaneIoU_{norm}(p_j, t_i)+\lambda f_{class}(p_j, t_i)
\end{equation}
where $\lambda$ is the parameter to balance the two costs and $f_{class}$ is the cost function for classification, such as a focal loss \cite{focalloss}. LaneIoU is normalized from its minimum to maximum.
The formulation (eq. \ref{eq:dynamick} and \ref{eq:cost}) realizes dynamic sample assignment of the proper number of anchors which are prioritized according to the evaluation metric. 
In summary, compared with CLRNet, our CLRerNet introduces LaneIoU as the dynamic-k assignment function, assignment cost function and loss function to learn the high-quality confidence scores that better correlate with the metric IoU.

\section{Experiments}
\subsection{Datasets}\label{subsec:datasets}

The CULane dataset\footnote{\url{https://xingangpan.github.io/projects/CULane.html}}\cite {pan2018SCNN} is the de-facto standard lane detection benchmark dataset which contains 88,880 train frames, 9,675 validation frames, and 34,680 test frames with lane point annotations. The test split has frame-based scene annotations such as Normal, Crowded and Curve (see Table \ref{table:sota_culane}). 
The CurveLanes\footnote{\url{https://github.com/SoulmateB/CurveLanes}} \cite{CurveLane-NAS} dataset contains the challenging curve scenes and consists of 100k train, 20k val and 30k test frames. We follow \cite{Liu_2021_ICCV} and use the val split for evaluation.

\noindent{\textbf{Removing the redundant train data.}} The CULane dataset includes a non-negligible amount of redundant frames where the ego-vehicle is stationary and the lane annotations do not change. We have found that overfitting to the redundant frames can be avoided by simply removing the frames whose average pixel value difference from the previous frame is below a threshold. The optimal threshold (=15) is chosen empirically via validation as described in the supplementary material. The remaining 55,698 (62.7\%) frames are utilized for training. The F1 score of CLRNet-DLA34 is improved from 80.30$\pm$0.05 to 80.86$\pm0.06$ ($N=5$ each) with the same 15-epoch training.

\subsection {Training and evaluation}\label{subsec:train_eval}
The models are implemented on PyTorch and MMDetection \cite{mmdetection}, and are trained for 15 epochs with AdamW \cite{loshchilov2018decoupled} optimizer. The initial learning rate is 0.0006 and cosine decay is applied. For CULane dataset, we crop the input image below $y=270$ and resize it to (800, 320) pixels. Horizontal flip, random brightness and contrast, random HSV modulation, motion and median blur and random affine modulations are adopted as data augmentation, following \cite{Zheng_2022_CVPR}. At the test time only the crop and resize are adopted and no test-time augmentations are applied.
In CLRerNet, LaneIoU is introduced as a loss function, dynamic-k calculation and assignment loss function. $w_{lane}$ is set to 15/800 for loss and dynamic-k, and 60/800 for cost to balance with the classification cost. The loss weights in eq. \ref{eq:lossses} are the same as \cite{Zheng_2022_CVPR} except for $\lambda_3$ which is set to 4. 
We additionally benchmark a CLRerNet-DLA34 model trained for 60 epochs applying exponential moving average (EMA). The learning rate decay is not applied and the momentum of EMA is set to 0.0001.

To validate the generality of our method, we add the LaneIoU-based sample assignment to LaneATT\cite{tabelini2021cvpr}.
Originally, LaneATT assigns non-learnable static anchors to GTs by horizontal distance thresholding.
We prioritize the anchors by calculating LaneIoU between predicted lanes and GTs to assign the positive-confidence targets.
More details are described in the supplementary material.

For CurveLanes \cite{CurveLane-NAS}, we follow the training setting of \cite{Liu_2021_ICCV} where the input resolution is (800, 320). To exploit the auxiliary segmentation loss, we draw the segmentation mask along all the lane labels with width of 30 pixels. Different from \cite{pan2018SCNN}, we set all the lane masks as class one (foreground). Since the test annotations are not available, we evaluate our method on the validation split. We employ the evaluation resolution of (224, 224) and line width of 5 following \cite{Liu_2021_ICCV}. $w_{lane}$ is set to 5/224 for loss and dynamic-k calculation and 20/224 for cost. $\lambda$ for assignment cost calculation (eq. \ref{eq:cost}) is set to 2.5. \\
\noindent{\textbf{Evaluation metric.}}
We employ F1 score \cite{pan2018SCNN} as an evaluation metric.
An IoU matrix between predicted lanes and ground-truths is calculated by comparing the segmentation masks drawn with a width of 30 pixels (Fig. \ref{fig1} bottom).
Based on the IoU matrix, one-to-one matching is calculated using linear sum assignment and the prediction-GT pairs with IoU over $t_{IoU}$ are considered as true positives ($TP$). Unmatched predictions and GTs are counted as false positives ($FP$) and false negatives ($FN$) respectively.
We employ two $t_{IoU}$ values for IoU calculation: 0.5 and 0.75. The F1 score is calculated as:
\begin{equation}
\label{F1score}
F_1=\frac{2\times Precision \times Recall}{Precision+Recall}
\end{equation}

\begin{figure}[t]
\begin{center}
   \includegraphics[width=0.99\linewidth]{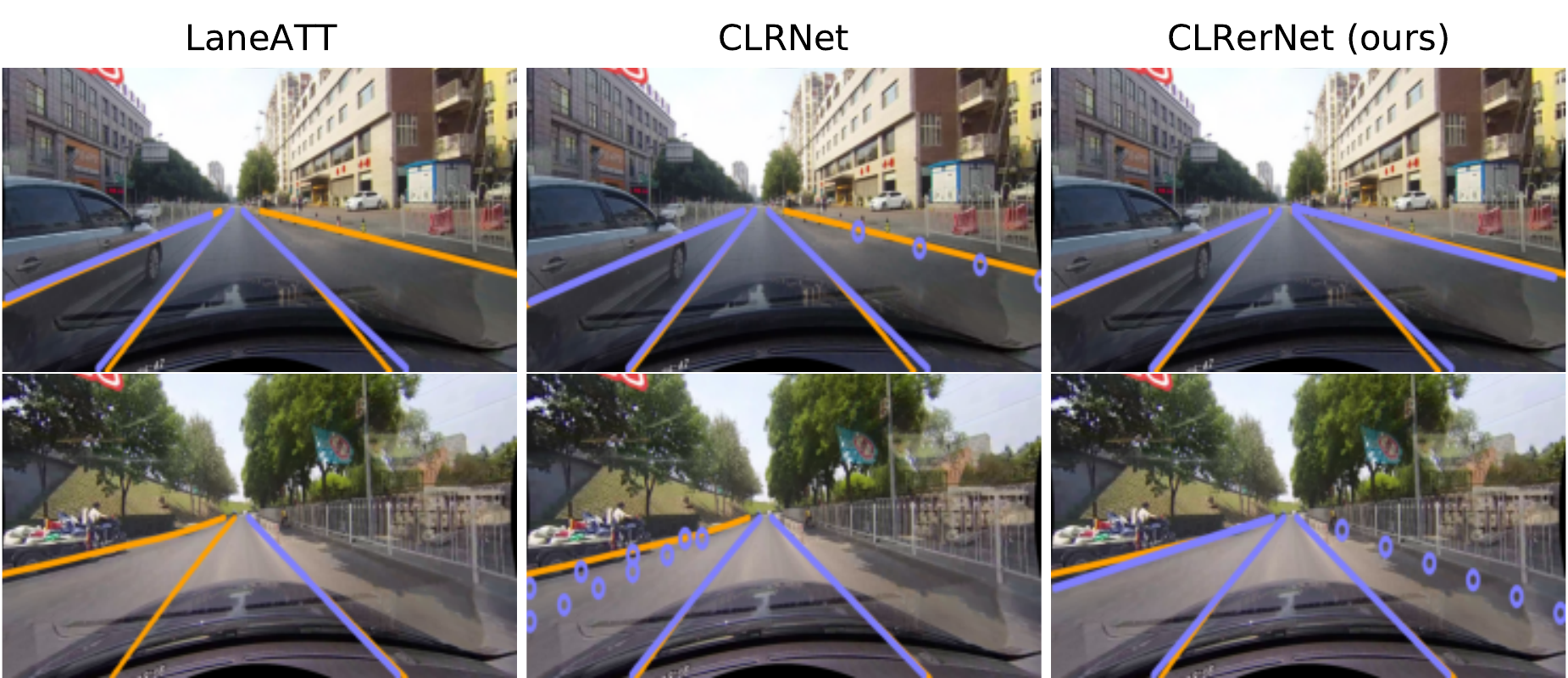}
  \end{center}
\caption{Qualitative results comparing LaneATT, CLRNet and our CLRerNet$\dagger\star$. Predictions and GTs are shown in blue and orange respectively. Predictions with insufficient confidence score are shown as blue circles.}
\label{fig:qualitative}
\end{figure}
\begin{table*}
\begin{adjustbox}{width=\textwidth} 
\begin{tabular}{ccccccccccccccc}
\hline
Method & Backbone & $\text{F1}_{50}$ &  $\text{F1}_{75}$  & Normal & Crowd & Dazzle & Shadow & Noline & Arrow& Curve  & Cross & Night & GFLOPs & FPS \\
\hline
SCNN & VGG16 & 71.60 & 39.84 & 90.60 & 69.70 & 58.50 & 66.90 & 43.40 & 84.10 & 64.40 & 1990 & 66.10 & 218.6 & 50 \\
LaneATT & Res18 & 75.13 & 51.29 & 91.17 & 72.71 & 65.82 & 68.03 & 49.13 & 87.82 & 63.75 & 1020 &  68.58 & 9.3 & 211 \\
LaneATT & Res34 & 76.68 & 54.34 &  92.14 & 75.03 & 66.47 & 78.15 & 49.39 & 88.38 & 67.72 & 1330 & 70.72 & 18.0 & 170 \\
CondLane\cite{Liu_2021_ICCV} & Res18 & 78.14 & 57.42 & 92.87 & 75.79 & 70.72 & 80.01 & 52.39 & 89.37 & 72.40 & 1364 & 73.23 & 10.2& 348\\
CondLane\cite{Liu_2021_ICCV}  & Res34 & 78.74 & 59.39 & 93.38 & 77.14 &  71.17 & 79.93 & 51.85 & 89.89 & 73.88 & 1387 & 73.92 & 19.6 & 237\\
CondLane\cite{Liu_2021_ICCV}  & Res101 & 79.48 & 61.23 & 93.47 & 77.44  & 70.93 & 80.91 & 54.13 & 90.16 & 75.21 & 1201& 74.80 & 44.8 & 97\\
CLRNet\cite{Zheng_2022_CVPR} & Res34 & 79.73 & 62.11 & 93.49 & 78.06 & 74.57 & 79.92 & 54.01 & 90.59 & 72.77 & 1216  & 75.02 & 21.5 &  204 \\
CLRNet\cite{Zheng_2022_CVPR} & Res101 & 80.13 & 62.96 & 93.85 & 78.78 & 72.49 & 82.33 & 54.50  & 89.79.& 75.57 & 1262 & 75.51 & 42.9 & 94 \\
CLRNet\cite{Zheng_2022_CVPR} & DLA34 &  80.47 & 62.78 & 93.73 & 79.59 & 75.30 & 82.51 & 54.58 & 90.62 & 74.13 & 1155 & 75.37 &  18.4 &  185 \\
\hline
\hline
LaneATT$\dagger$ & Res34 & 77.51\footnotesize{$\pm$0.10} &56.78 & 92.48 & 75.47 & 68.09 & 73.21 & 50.96 & 88.72 & 68.18 & 1054 & 72.58 & 18.0 & 170 \\
CLRNet$\dagger$  & Res34 & 80.54\footnotesize{$\pm$0.12} &63.65 & 93.85 & 79.22 & 73.32 & 82.50 & 55.26 & 90.84 & 74.06 & 1106 & 75.92 &  21.5 & 204 \\
CLRNet$\dagger$  & Res101 &80.67\footnotesize{$\pm$0.06} &64.35 & 93.95 & 79.60 & 72.91 & 81.58 & 55.76 & 90.42 & 74.06 & 1166 & 76.01 & 42.9 & 94 \\
CLRNet$\dagger$  & DLA34 & 80.86\footnotesize{$\pm$0.06} &64.05 & 94.03 & 79.78 & 75.23 & 81.94 & 56.02 & 90.67 & 74.57 & 1184 & 76.40 & 18.4 &  185 \\
\hline
LaneATT+$\dagger$ & Res34 &78.19\footnotesize{$\pm$0.06} &56.96 & 92.60 & 76.42 & 69.12 & 77.59 & 52.01 & 88.75 & 64.49 & \textbf{974} & 72.78 & 18.0 & 153 \\
CLRerNet$\dagger$ & Res34 & 80.76\footnotesize{$\pm$0.13} &63.77 & 93.93 & 79.51 & 73.88 & 83.16 & 55.55 & 90.87 & 74.45 & 1088 & 76.02 &  21.5 &  204 \\
CLRerNet$\dagger$ & Res101 & 80.91\footnotesize{$\pm$0.10} &64.30 & 93.91 & 80.03 & 72.98 & 82.92 & 55.73 & 90.53 & 73.83 & 1113 & 76.13 &   42.9 & 94 \\
CLRerNet$\dagger$ & DLA34 & 81.12\footnotesize{$\pm$0.04} & 64.07 & 94.02 & 80.20 & 74.41 & 83.71 & 56.27 & 90.39 & 74.67 & 1161 & 76.53 &    18.4 & 185 \\
\hline
CLRerNet$\dagger\star$ & DLA34 & \textbf{81.43}\footnotesize{$\pm$0.14} & \textbf{65.06} & \textbf{94.36} & \textbf{80.62} & \textbf{75.23} & \textbf{84.35} & \textbf{57.31} & \textbf{91.17} & \textbf{79.11} & 1540 & \textbf{76.92} &    18.4 & 185 \\
\hline
\end{tabular}
\end{adjustbox}
\caption{Evaluation results on the CULane test set. Our experiments are below the double horizontal line.}
\label{table:sota_culane}
\end{table*}

\noindent{\textbf{Cross validation and test.}} 
The F1 metric is sensitive to the threshold of detection confidence score. 
We perform 5-fold cross validation on the train split, by randomly dividing the train videos into five groups. 
The F1 score of each threshold is averaged across the 5-fold results, and the optimal threshold is determined by taking the argmax of it.

Moreover, we find that the F1 score deviation of the models trained with different random seeds is not negligible.
For instance, the F1 score of the CLRNet-DLA34 ranges from the minimum of 80.20 to the maximum of 80.34 ($N=5$). 
For a more reliable and fairer benchmark on the test split, we train five models with different random seeds for each condition and calculate the mean and standard deviation of the metrics, on which the confidence thresholds obtained by five-fold cross validation is applied.
To the best of our knowledge, we are the first to conduct the above benchmark protocol in lane detection.
As for the CurveLanes dataset where the test annotations are not available, we report the maximum F1 score on the validation split with respect to the confidence thresholds.

\subsection{CULane benchmark results}
The benchmark results on the CULane test set are shown in Table \ref{table:sota_culane}. 
The rows below the double horizontal line are our experiment results. 
For each condition (row), we show the averaged metric values of five models trained with different seeds.
The confidence threshold obtained from 5-fold cross-validation is employed. 
The $\text{F1}_{50}$ scores are shown in the test scene columns except for the \textit{cross} metric where the number of false positives is shown.
All the FPS results on Table \ref{table:sota_culane} are measured with a GeForce RTX 3090 GPU.
CLRNet$\dagger$ and LaneATT$\dagger$ are the baseline model trained with our implementation.
Our method CLRerNet$\dagger$ employs LaneIoU for dynamic-k calculation, assignment cost and loss functions. CLRerNet$\dagger\star$ is the boosted version of CLRerNet$\dagger$ which is trained for 60 epochs with EMA.

With introducing LaneIoU, CLRerNet$\dagger$ with DLA34 outperforms CLRNet$\dagger$ by 0.26\% in $\text{F1}_{50}$. Moreover, the boosted model CLRerNet$\dagger\star$ reaches $\text{F1}_{50}=81.43\%$ in average, enjoying the state-of-the-art performance surpassing the previous methods ($\text{F1}_{50}=80.47\%$, single experiment of CLRNet+DLA34) by a large margin.
The performance improvement by LaneIoU is also observed on the models with other backbones - 80.54\% to 80.76\% ($+0.22\%$) with ResNet34 and 80.67\% to 80.91\% ($+0.24\%$) with ResNet101.
LaneATT+$\dagger$ is improved by the LaneIoU-based assignment by 0.68\%, validating the generality of our method.
CLRerNet does not increase test-time computational complexity and shows the same GFLOPs and FPS as CLRNet.

Qualitative results on the CULane test set are shown in Fig. \ref{fig:qualitative}. Our CLRerNet$\dagger\star$ is capable of detecting the lanes in the challenging scenes. 
The right-most tilted lane of the first image (top) and the left-most lane of the second image (bottom) are detected only by CLRerNet with high confidence scores.
The examples qualitatively suggest that CLRerNet is able to give more correct scores to predictions, which is analyzed in Subsection \ref{ablation}.

\subsection{CurveLanes validation results}
The validation results on CurveLanes are shown in Table \ref{table:curvelanes}.
The default CLRNet \cite{Zheng_2022_CVPR} with the DLA34 backbone shows the same F1 score as CondLane-L\cite{Liu_2021_ICCV} with lower computation cost. Note that our results are the average of five training trials. The confidence threshold is set to the empirically optimal value 0.44. CLRerNet significantly outperforms the baseline by 0.37\%, achieving the new state-of-the-art 86.47\%.

\begin{table}
\begin{center}
\begin{adjustbox}{width=\linewidth} 
\begin{tabular}{ccccc}
\hline
Method & F1 & Precision & Recall & GFLOPs \\
\hline
CondLane-S\cite{Liu_2021_ICCV}  & 85.09 & 87.75 & 82.58 & 10.3 \\
CondLane-M\cite{Liu_2021_ICCV} & 85.92 & 88.29 & 83.68 & 19.7 \\
CondLane-L\cite{Liu_2021_ICCV} & 86.10 & 88.98 & 83.41  & 44.9 \\
\hline\hline
CLRNet-DLA34\cite{Zheng_2022_CVPR} & 86.10\footnotesize{$\pm$0.08 }& 91.40 & 81.39  & 18.4 \\
CLRerNet-DLA34 & \textbf{86.47}\footnotesize{$\pm$0.07} & 91.66 & 81.83 & 18.4 \\
\hline
\end{tabular}
\end{adjustbox}
\end{center}
\caption{Comparison between methods on the CurveLanes \textit{val} set. Our experiments are below the double horizontal line.}
\label{table:curvelanes}
\end{table}

\subsection{Ablation study and analysis}\label{ablation}
We corroborate the effectiveness of our method by ablating LaneIoU from dynamic-k calculation, assignment cost and loss function.
CLRerNet with DLA34 backbone is trained in each condition with the redundant train data omitted.
We follow the benchmark protocol described in subsection \ref{subsec:train_eval}, thus ten models (5 seeds + 5 folds) are trained and validated for each condition. 
The results in Table \ref{table:ablation} show that the performance degrades by replacing LaneIoU with \cite{Zheng_2022_CVPR} for dynamic-k determination, cost function and loss function respectively.
Determining the number of assignments each GT lane by LaneIoU mitigates the inhomogeneity caused by lane tilt variation.
The LaneIoU-based assignment cost (eq. \ref{eq:cost}) prioritizes the predicted lanes which have higher metric IoU with the GTs, leading to more accurate confidence learning as motivated in subsection \ref{subsection:oracle}.
Replacing LineIoU loss with LaneIoU loss also mitigates the tilt dependency of the regression penalty.

\begin{figure}[t]
\begin{center}
 \includegraphics[width=.9\linewidth]{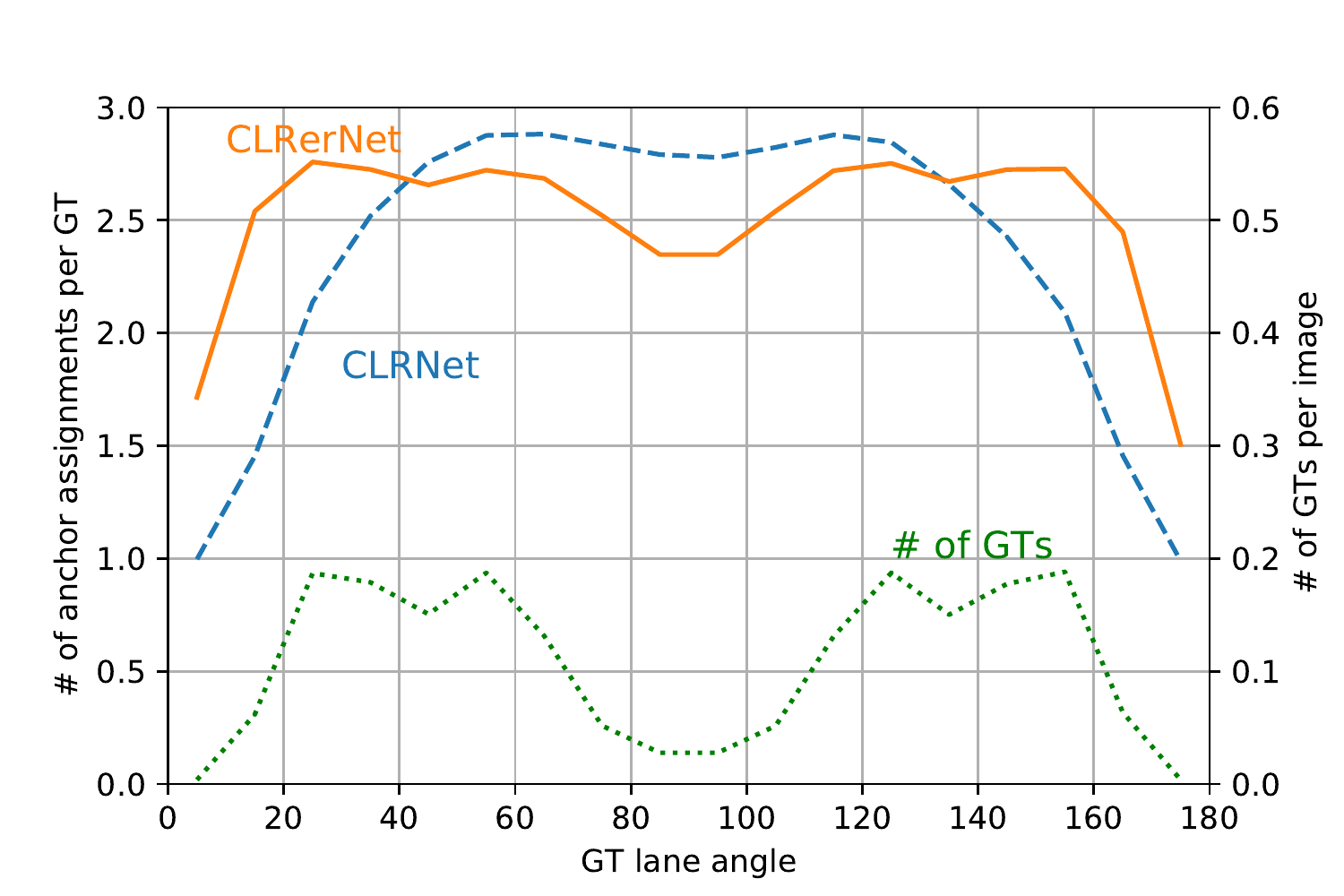}
\end{center}
   \caption{The average number of assignments in different angle ranges.}
\label{fig:angle_dependency}
\end{figure}

Fig. \ref{fig:angle_dependency} shows the comparison between CLRerNet and CLRNet in terms of anchor assignment numbers per GT in different angle ranges.
The assignment numbers are accumulated during the training and averaged.
The angles are calculated using GT lanes in (800, 320) resolution and $90^{\circ}$ corresponds to the vertical lane.
By leveraging LaneIoU, the assignment number becomes more homogeneous with respect to the lane angles, especially in the angle ranges of $20^{\circ}$ to $60^{\circ}$ and $120^{\circ}$ to $160^{\circ}$ where the GTs typically exist.

\begin{figure}[t]
\begin{center}
 \includegraphics[width=.9\linewidth]{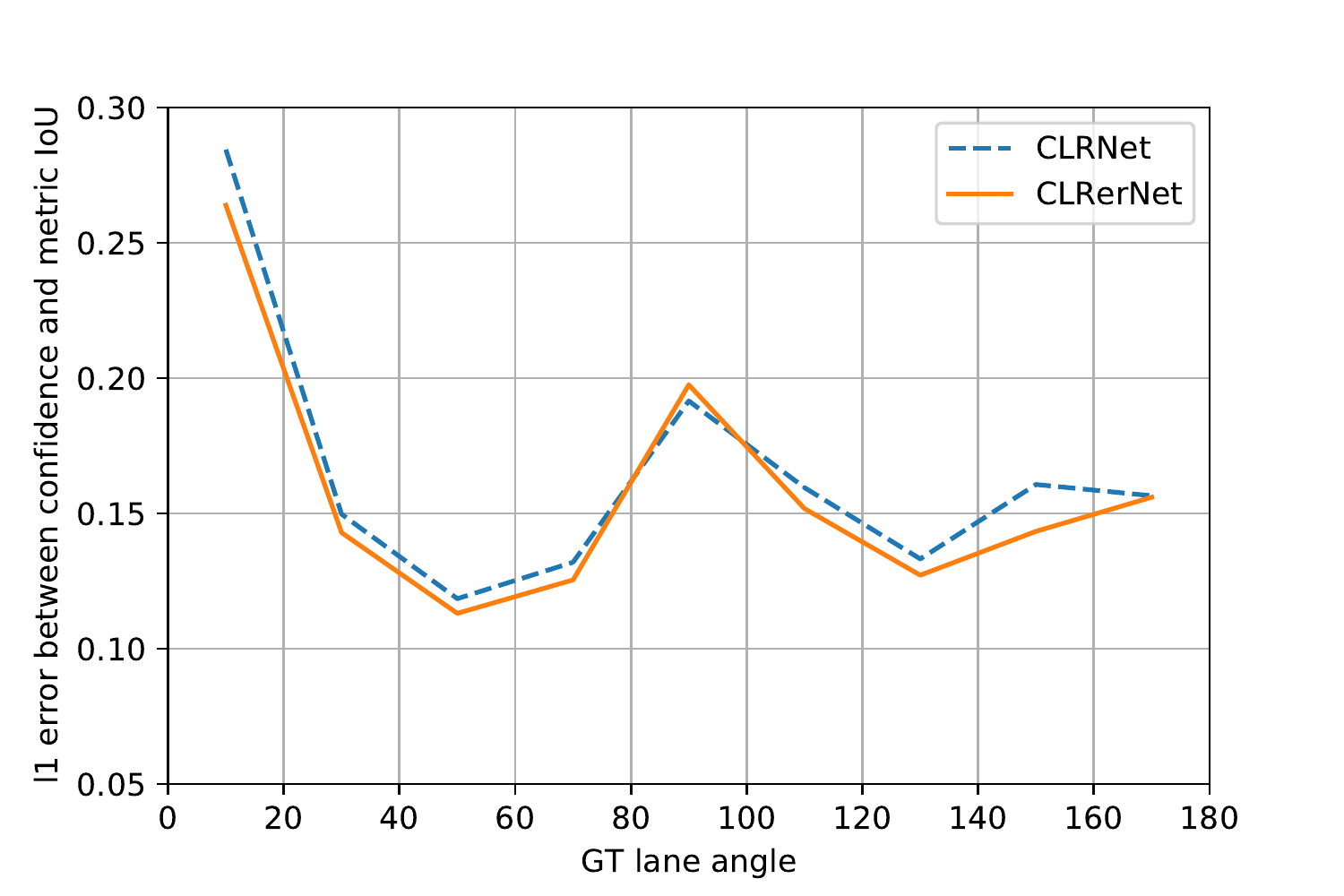}
\end{center}
   \caption{The average $l1$ error between confidence score and metric IoU in different angle ranges.}
\label{fig:angle_l1}
\end{figure}

The assigned anchor's confidence target is set to positive prioritized by LaneIoU.
Therefore, the confidence is trained more homogeneously across different lane angles.
As can be seen in Fig. \ref{fig:angle_l1}, the $l1$ error between the predicted confidence scores and the metric IoU values is improved in CLRerNet in the non-vertical angle ranges, corroborating the effectiveness of LaneIoU.

\begin{table}
\begin{center}
\begin{adjustbox}{width=\linewidth} 
\begin{tabular}{cccccc}
\hline
dynamic-k & cost  & loss & $\text{F1}_{50}$ &  $\text{F1}_{75}$  \\
\hline
\cite{Zheng_2022_CVPR}  & \cite{Zheng_2022_CVPR}  &  \cite{Zheng_2022_CVPR}  & 80.86\footnotesize{$\pm$0.06} &64.05\footnotesize{$\pm$0.17} \\
LaneIoU & \cite{Zheng_2022_CVPR} & \cite{Zheng_2022_CVPR}  & 80.98\footnotesize{$\pm$0.07} &64.17\footnotesize{$\pm$0.17}  \\
LaneIoU & LaneIoU & \cite{Zheng_2022_CVPR}  & 81.07\footnotesize{$\pm$0.03} &64.22\footnotesize{$\pm$0.26}  \\
LaneIoU & LaneIoU  & LaneIoU & \textbf{81.12}\footnotesize{$\pm$0.05} & \textbf{64.28}\footnotesize{$\pm$0.15}  \\
\hline
\end{tabular}
\end{adjustbox}
\end{center}
\caption{Ablation study by replacing LaneIoU (ours) with \cite{Zheng_2022_CVPR}.}
\label{table:ablation}
\end{table}

\begin{figure}[t]
\begin{center}
   \includegraphics[width=0.99\linewidth]{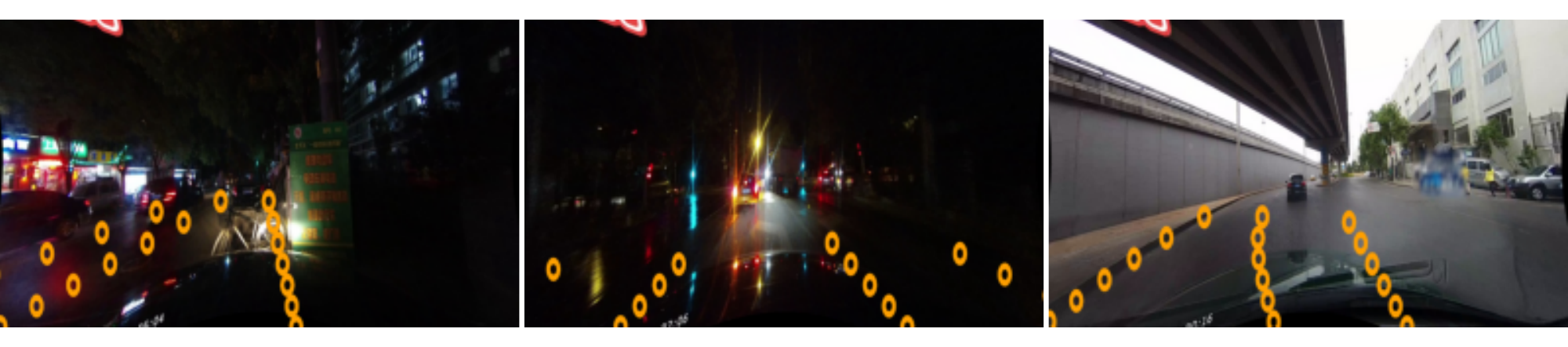}
  \end{center}
\caption{Extremely difficult cases in the CULane dataset. GTs are overlaid as orange circles.}
\label{fig:qualitative_extreme}
\end{figure}

\noindent{\textbf{Discussion.}} Although CLRerNet shows significant improvement in performance, there still is a gap between the best CLRerNet model's performance (81.43\%) and the oracle-confidence case (98.47\%). 
The dataset-oriented issues including label fluctuation and data imbalance are considered to be the part of the gap. 
For instance, there are the cases in the CULane test set where detection is extremely difficult (Fig. \ref{fig:qualitative_extreme}). 
As can be found in Table \ref{table:sota_culane}, the \textit{Noline} test category is the most challenging as there are no visual markings on the road. Such cases are prone to label fluctuation and inconsistency. 
Likewise, data imbalance such as stationary scenes greatly affects the model training. As is mentioned in Subsection \ref{subsec:datasets}, we find that mitigating the data imbalance significantly improves the performance. 
\section{Conclusion}

We disentangle the lane prediction components by the oracle experiment and demonstrate the importance of high-quality confidence scores for more accurate lane detection.
To make confidence scores represent the metric IoU, the novel LaneIoU is proposed and integrated into the row-based lane detection baselines.
A novel detector coined CLRerNet is developed by introducing LaneIoU as the sample assignment and loss functions.
The statistical and fair benchmark protocol is employed utilizing five-seed models and five-fold cross validation.
CLRerNet achieves the state-of-the-art performance on the challenging CULane and CurveLanes datasets significantly surpassing the baseline.
We believe our oracle experiments, LaneIoU-based training and benchmark protocol bring a clearer view of lane detection to the community.
\clearpage
{\small
\bibliographystyle{ieee_fullname}
\bibliography{lane}
}

\end{document}